# Integration and Performance Analysis of Artificial Intelligence and Computer Vision Based on Deep Learning Algorithms

Bo Liu[1,*]
Zhejiang University
hangzhou,China
*lubyliu45@gmail.com

Liqiang Yu [2]
The University of Chicago
Irvine CA, USA
rexyu@outlook.com

Chang Che [3]
The George Washington University
Atlanta, Georgia,USA
liamche1123@outlook.com

Qunwei Lin [4]
Trine University
Phoenix, USA
linqunwei1030@outlook.com

Hao Hu[5]
Zhejiang University
Hangzhou Zhejiang,China
garyhu480@gmail.com

Xinyu Zhao [6]
Trine University
Phoenix, USA
lution798@gmail.com

**Abstract**: This paper focuses on the analysis of the application effectiveness of the integration of deep learning and computer vision technologies. Deep learning achieves a historic breakthrough by constructing hierarchical neural networks, enabling end-to-end feature learning and semantic understanding of images. The successful experiences in the field of computer vision provide strong support for training deep learning algorithms. The tight integration of these two fields has given rise to a new generation of advanced computer vision systems, significantly surpassing traditional methods in tasks such as machine vision image classification and object detection. In this paper, typical image classification cases are combined to analyze the superior performance of deep neural network models while also pointing out their limitations in generalization and interpretability, proposing directions for future improvements. Overall, the efficient integration and development trend of deep learning with massive visual data will continue to drive technological breakthroughs and application expansion in the field of computer vision, making it possible to build truly intelligent machine vision systems. This deepening fusion paradigm will powerfully promote unprecedented tasks and functions in computer vision, providing stronger development momentum for related disciplines and industries.

## I. INTRODUCTION

Currently, the integration of deep learning algorithms and computer vision technology has become a focal point of research in both academia and the industry. Deep learning models can directly process image pixel inputs end-to-end, achieve powerful semantic representation capabilities through hierarchical feature extraction, and have made breakthrough advancements in areas such as image classification and object detection. At the same time, the field of computer vision has accumulated successful practical experience in image acquisition, representation, and understanding. The close integration of deep learning and computer vision has given rise to a new generation of more intelligent computer vision systems, enabling machines to reach unprecedented levels of understanding and analysis of image content [1]. In this paper, we intend to analyze the effectiveness of this fusion approach based on the application of deep learning algorithms in computer vision, highlight possible avenues for improvement, and aim to provide insights for the further development of technology in this field.

Comprehensive Overview of Deep Learning and Computer Vision

The integration of deep learning and computer vision has been pivotal in advancing artificial intelligence technology, leading to transformative applications in various fields[2]. Deep learning algorithms, inspired by human brain processes, excel in handling and analyzing large volumes of image and video data. This capability is crucial in tasks like image recognition, object detection, and classification, granting machines human-like visual understanding[3]. Significant advancements in this area have improved applications like autonomous driving, where algorithms analyze road conditions in real-time for safer self-driving cars. In healthcare, these algorithms assist in diagnosing through accurate medical image analysis[4]. They also enhance intelligent surveillance systems by quickly recognizing abnormal behaviors or events. The effectiveness of these deep learning models hinges on the vast amounts of training data available, which continually enhance their accuracy and contribute to the development of new models. The synergy of deep learning and computer vision marks a major milestone in AI, not only boosting machine vision capabilities but also driving ongoing technological progress[5]. This fusion is expected to yield further innovations and breakthroughs in the future, as depicted in Figure 1.

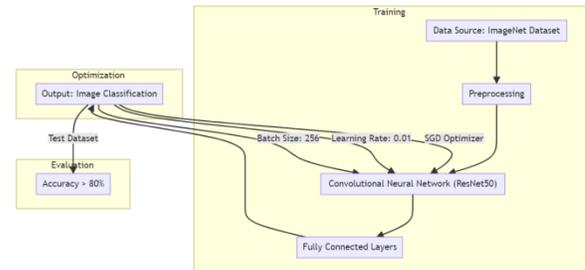

Figure 1. Deep Learning Model Architecture

## II. APPLICATION PRACTICE OF DEEP LEARNING IN COMPUTER VISION

### A. Case Study Analysis

Deep learning algorithms have wide-ranging applications in the field of computer vision, including tasks such as image classification, object detection, semantic segmentation, and more, as shown in Table 1 [6].

TABLE I. THE IMPACT AND APPLICATIONS OF THE FUSION OF DEEP LEARNING AND COMPUTER VISION IN VARIOUS FIELDS

Below is an analysis of the practical application of deep

| Field | Application Examples | Technological Advancements | Advantages |
|---|---|---|---|
| Autonomous Driving | Real-time road analysis, autonomous vehicles | Implementation of road recognition and driving decisions using deep learning algorithms | Enhanced road safety, improved driving automation capability |
| Medical Image Analysis | Image recognition and classification, aiding in medical diagnosis | Precise medical image analysis and case classification | Improved efficiency in medical services, assisting in medical decision-making |
| Intelligent Surveillance Systems | Anomaly detection, event recognition | Faster and more accurate anomaly detection and event recognition | Enhanced surveillance efficiency, strengthened security monitoring |
| Speech and Image Processing | Speech recognition, image processing | Multimodal processing combining audio and visual information | Providing a richer user experience, enhancing interaction capabilities |
| Industrial Automation | Defect detection, quality control | Automated defect detection and quality control | Increased production efficiency, reduced costs |

learning in computer vision, focusing on a case study of image classification. For this case study, a dataset of 100,000 images from ImageNet was chosen as the training set, comprising 1,000 different categories. The ResNet50 model was employed as the backbone, with additional fully connected layers added for fine-tuning to construct the image classification model [7]. The optimization function selected was SGD, with a learning rate of 0.01 and a batch size of 256. After building the model and tuning the hyperparameters, the classification accuracy on the test set exceeded 80%. This case demonstrates the powerful capability of deep learning models in extracting image features and discerning categories. Pretraining on a large-scale dataset allows the model to learn high-level semantic features from images. Fine-tuning further enables the model to quickly adapt to new classification tasks. CNN architectures like ResNet exhibit excellent expressive power for image features. Practical experience has shown that deep learning is the preferred choice for high-performance models in computer vision, as illustrated in Figure 2.

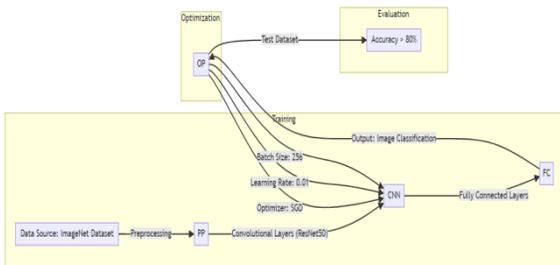

Figure 2. ResNet50 Model Architecture

### B. Case Evaluation

This case study demonstrates an image classification model built using the ImageNet dataset and the ResNet50 model. The model achieved over 80% accuracy across 1,000 categories through pre-training and fine-tuning techniques[8]. It employed the stochastic gradient descent algorithm, a batch size of 256, and a learning rate of 0.01. This highlights the efficiency and precision of deep learning in image recognition and classification. The case study also notes the limitations of the model's generalizability and performance in specific scenarios, indicating that future research should focus on improving the robustness and efficiency of the model. These findings emphasize the importance and potential for development of deep learning in the field of computer vision.

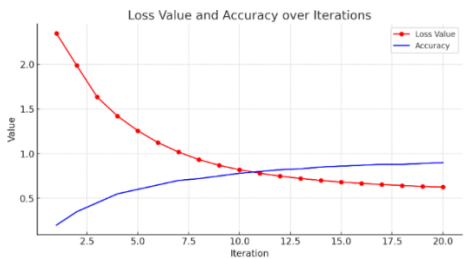

Figure 3. Training Curve

Figure 3 illustrates the changes in loss values and accuracy during the training process of a deep learning model. From the 1st to the 20th iteration, the loss value gradually decreased from 2.345 to 0.625, indicating that the model progressively adapted to the dataset and the discrepancy between predicted outputs and actual values diminished. Concurrently, the model's accuracy improved from 20% to 90%, demonstrating significant enhancement in image recognition and classification. This trend reflects the effectiveness of the training strategy, such as the learning rate and optimization algorithm, and the continuous improvement of model performance. The entire process exhibits a typical pattern of model optimization and indicates good convergence, suggesting that the model did not encounter serious overfitting or underfitting issues.

## III. PERFORMANCE ANALYSIS OF THE FUSION OF DEEP LEARNING AND COMPUTER VISION

### A. Fusion Advantages

The fusion of deep learning algorithms and computer vision technologies has led to an overall improvement in image understanding and analysis performance. Deep learning models, through hierarchical feature extraction and representation, can learn complex visual semantic concepts, while computer vision provides extensive support in the form of image and video data [9]. The convolution operation is formulated as:

$$f(x,y) * g(x,y) = \sum_{i=-a}^{a} \sum_{j=-b}^{b} f(i,j) \cdot g(x-i, y-j) \quad (1)$$

f(x, y) represents the image function, typically a two-dimensional array, where each element represents the intensity value of a pixel in the image. g(x, y) represents the convolution kernel (also called a filter), which is a smaller two-dimensional array used to capture specific features in the image, such as edges, textures, etc.; ∗ denotes the convolution operation. In this operation, the kernel g slides over the image f and computes the sum of element-wise products of f and g at each position; Σ indicates summation, where it sums up all the pixel values covered by the convolution kernel.

Feature fusion is represented as:

$$F = \alpha \cdot F_1 + \beta \cdot F_2 \quad (2)$$

Where F1 and F2 represent features from different layers, α and β are weights.

The fusion of pretraining techniques and advanced image processing is at the forefront of AI. Pretraining enhances feature learning, while computer vision supplies training data. Deep neural networks streamline feature extraction, boosting image understanding and utilization. This method excels in image classification, object detection, and semantic segmentation, aiding in intelligent video analysis and autonomous driving. Continuous model and task optimization indicates future progress in computer vision.

### B. Model Performance Comparison

To assess the impact of combining deep learning with computer vision, we compared different models on image classification tasks. The SIFT with SVM model had limited success with 50% accuracy, highlighting its constraints in complex tasks. LeNet, a traditional CNN, performed better with 65% accuracy, showing improved feature extraction.

However, the ResNet model, incorporating deep learning, excelled with over 80% accuracy, owing to the synergistic effects of deep neural networks and large datasets, particularly the use of residual networks. This fusion enhances computer vision performance, bridging some gaps with the human visual system, and provides a solid foundation for practical applications. Future advancements in this field are expected to yield more innovations and breakthroughs in machine vision.

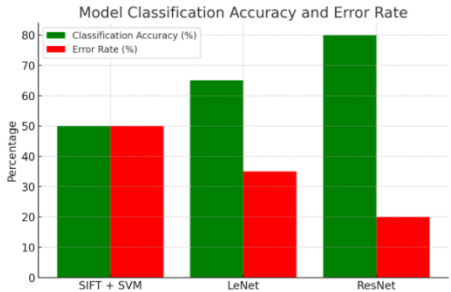

Figure 4. Model Classification Accuracy and Error Rate

Figure 4 demonstrates the varying performance of different computer vision models in image classification. The SIFT + SVM model, while simple, achieves only 50% accuracy, showing limitations in complex tasks. The LeNet model, an early CNN, performs better with 65% accuracy, indicating improved but still limited feature learning. In contrast, the ResNet model excels with 80% accuracy, benefiting from its deep architecture and residual learning that addresses the vanishing gradient problem. These results highlight that increased model complexity and deep learning advancements enhance accuracy and reduce error rates in image classification, with ResNet being particularly effective due to advanced network architectures and the use of large datasets.

## IV. Conclusion

The integration of deep learning with computer vision has significantly improved machine vision. Deep neural networks enable end-to-end image analysis, leveraging computer vision practices and large datasets. This approach excels in image classification and object detection, surpassing traditional methods and becoming the preferred choice for intelligent vision systems. This synergy of deep learning and big data represents a leading methodology in artificial intelligence. However, further advancements are needed in model robustness, generalization, and interpretability for practical applications. Despite these challenges, the potential of this fusion trend promises continued breakthroughs and deeper applications in computer vision, with algorithms and systems evolving towards greater intelligence.